\PassOptionsToPackage{table}{xcolor}
\documentclass[10pt,twocolumn,letterpaper]{article}

\usepackage[pagenumbers]{cvpr} 

\usepackage[utf8]{inputenc} 
\usepackage[T1]{fontenc}    

\usepackage{url}            
\usepackage{booktabs}       
\usepackage{amsfonts}       
\usepackage{nicefrac}       
\usepackage{microtype}      
\usepackage{xcolor}         
\usepackage{float}

\usepackage{array} 
\usepackage{multirow}
\usepackage{eqnarray}
\usepackage{amsmath}
\usepackage{graphicx}
\graphicspath{ {./Figs/} }
\usepackage{caption}
\usepackage{subcaption}
\usepackage{svg}
\usepackage{enumitem}

\usepackage{rotating}

\usepackage{amssymb}
\usepackage{pifont}

\usepackage[accsupp]{axessibility}

\usepackage[outdir=./Figs/]{epstopdf}

\newcommand{\shortand}{\hspace{17pt}}

\definecolor{cvprblue}{rgb}{0.21,0.49,0.74}
\usepackage[pagebackref,breaklinks,colorlinks,citecolor=cvprblue]{hyperref}

\newcommand{\Yesred}{{ \color{red} Yes }}
\definecolor{darkgreen}{rgb}{0.0, 0.5, 0.0}
\definecolor{moderategrey}{rgb}{0.8, 0.8, 0.8} 

\newcommand{\Nogreen}{{\color{darkgreen} No}}

\def\paperID{15890} 
\def\confName{CVPR}
\def\confYear{2025}

\title{Learning Physics From Video: \\ Unsupervised Physical Parameter Estimation for Continuous Dynamical Systems}

\author{Alejandro Castañeda Garcia \shortand Jan Warchocki \shortand Jan van Gemert\shortand Daan Brinks\shortand Nergis Tömen\\
Delft University of Technology\\
}

\begin{document}
\maketitle
\begin{abstract}

Extracting physical dynamical system parameters from recorded observations is key in natural science. Current methods for automatic  parameter estimation from video train supervised deep networks on large datasets. Such datasets require labels, which are difficult to acquire. While some unsupervised techniques--which depend on frame prediction--exist, they suffer from long training times, initialization instabilities, only consider motion-based dynamical systems, and are evaluated mainly on synthetic data. In this work, we propose an unsupervised method to estimate the physical parameters of known, continuous governing equations from single videos suitable for different dynamical systems beyond motion and robust to initialization. Moreover, we remove the need for frame prediction by implementing a KL-divergence-based loss function in the latent space, which avoids convergence to trivial solutions and reduces model size and compute. We first evaluate our model on  synthetic data, as commonly done. After which, we take the field closer to reality by recording Delfys75: our own real-world dataset  of 75 videos for five different types of dynamical systems to evaluate our method and others. Our method compares favorably to others.  
Code and data are available online:~\url{https://github.com/Alejandro-neuro/Learning_physics_from_video}. 
\end{abstract}    
\vspace{-0.1cm}
\section{Introduction}

Estimating dynamical parameters of physical and biological systems from videos allows relating visual data to known governing equations which can be used to make predictions, improve mathematical models, understand diseases, and, in general, advance our knowledge in science and technology~\citep{brunton2016discovering,jaques2019physics,schmidt2009distilling}. Use cases include trajectory prediction for celestial objects~\citep{hofherr2023neural}, healthy and diseased tissue characterization ~\cite{gutenkunst2007universally}, and physical model validation~\citep{brunton2016discovering, champion2019data}.
\begin{figure}[t] 
    \centering
    \includegraphics[width=0.45\textwidth]{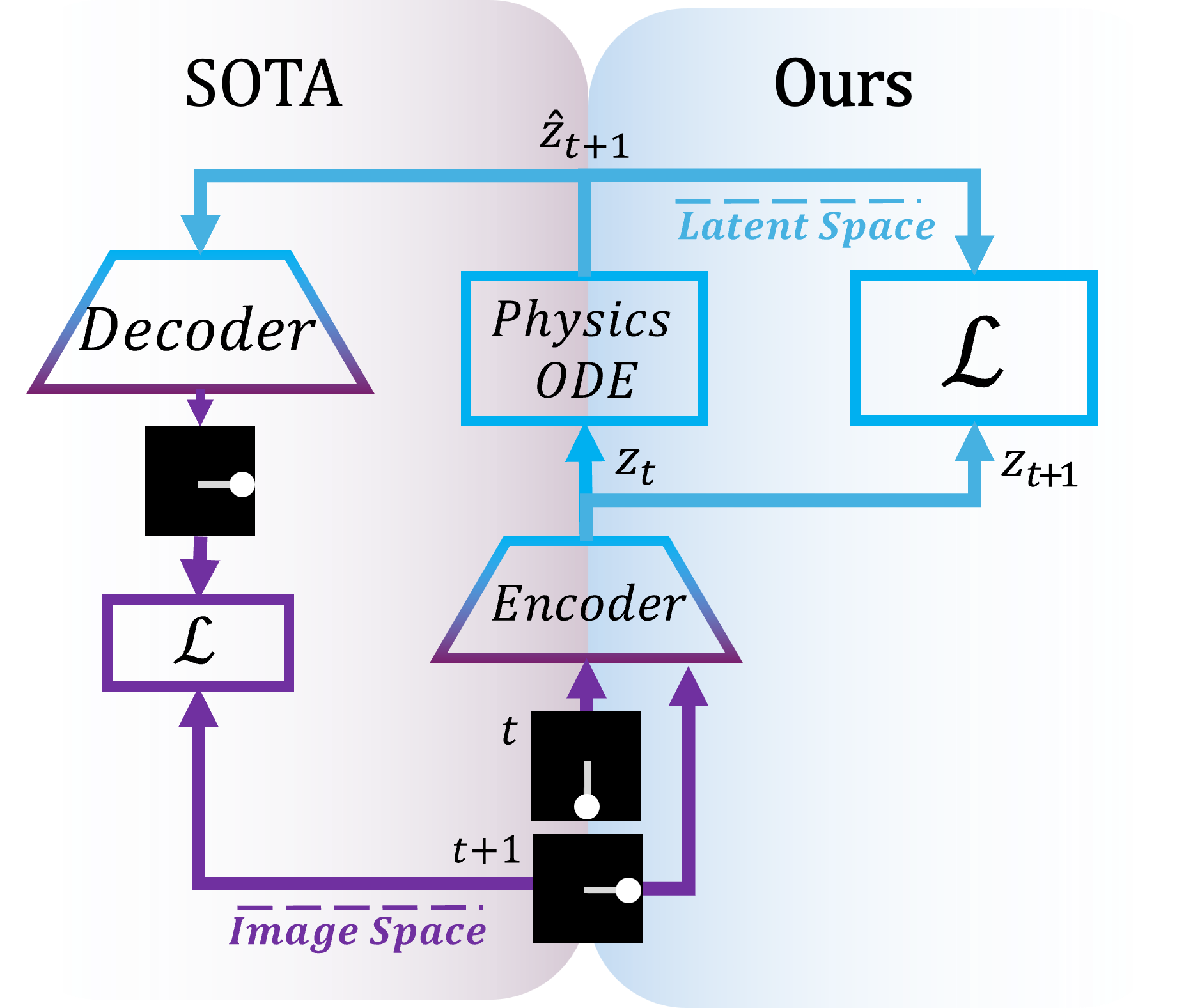}
    \caption{We propose a novel unsupervised approach to physical parameter estimation from videos. Black squares are video frames with different states of a white pendulum. Starting from a frame at time $t$ (center) an encoder estimates the dynamical states $z_t$. A learnable physics block (Physics ODE) solves the dynamical system equations to predict future states $\hat{z}_{t+1}$ in latent space (blue lines). Previous state-of-the-art methods (left) then use decoders and a reconstruction loss ($\mathcal{L}$, purple left) to train the physics ODE block. In contrast, our method (right) completely avoids the need for a decoder by leveraging a loss function in the latent space ($\mathcal{L}$, blue right). Our loss function minimizes the distance between the estimated states $\hat{z}_{t+1}$ and $z_{t+1}$.}
    \label{fig:Figure1}
    \vspace{-0.2cm}
\end{figure}

Fitting governing equations is an inverse problem~\citep{anderson2015chapter}, which often requires using additional sensors to directly measure system states. Video-based measurements can eliminate the need for additional sensors, yet, require manually labelling pixels or video frames which is time-consuming and expensive. Therefore, automated and unsupervised methods are key to extract dynamics from videos and accurately estimate physical parameters~\citep{brunton2016discovering, hofherr2023neural, jaques2019physics, iten2020discovering, schmidt2009distilling}. 

Recent work addressed parameter estimation from video by deep learning~\citep{brunton2016discovering, champion2019data,watters2017visual} or reinforcement learning~\citep{asenov2019vid2param}. Supervised methods rely on datasets with extensive and high precision labels which are exceedingly difficult to obtain~\citep{adam2015higgs, ball2010data,  libbrecht2015machine, meijering2016imagining, varoquaux2022machine}. To avoid labeling, current unsupervised methods for estimating physical parameters build on encoder-decoder network designs: reconstructing video frames from low-dimensional representations. However, frame reconstruction is a mere by-product of the parameter estimation and leads to hardwired architectures which cannot be extended to different dynamics~\cite {hofherr2023neural, jaques2019physics}. Consequently, current solutions~\citep{hofherr2023neural, jaques2019physics, kandukuri2020learning, weiss2020correspondence, NEURIPS2020_79f56e5e} are constrained to motion dynamics, excluding a wide variety of systems with dynamics related to brightness, color, and deformations, among others~\citep{hofherr2023neural,jaques2019physics}. While such methods have been shown to work well on synthetic data~\cite{jaques2019physics, hofherr2023neural, yang2022learning}, they are resource intensive, sensitive to initialization, and it is not clear how well they will perform on realistic data.

Here, we propose an unsupervised method to solve the inverse problem using videos of dynamical systems governed by known continuous equations. Unlike prior approaches, our method is versatile across various systems beyond motion. We use 
a loss in latent space that eliminates the need for a reconstruction decoder. This approach is faster, less resource-intensive, and more robust to initial conditions than existing methods.  Additionally, to take a step towards real-world applications,  we collect the Delfys75 dataset: 75 real-world videos across five dynamical systems, with ground-truth parameters for motion, brightness, and scale dynamics. We benchmark baselines and our model, where we compare favorably. The proposed method is visualised in Figure~\ref{fig:Figure1}. Our main contributions are summarized as:

\begin{itemize}[left=0.1cm, itemsep=0.1em, labelsep=0.1cm]
    \item Accurate dynamical state estimation from video with precise extrapolation at test time.
    \item A decoder-free model, capable of modeling dynamics beyond motion, robust to changes in initial conditions.
    \item An unsupervised latent space loss which avoids model collapse in unsupervised parameter estimation.
    \item Delfys75: a real-world dataset of 75 videos for 5 different physical systems with annotated ground truth. 
\end{itemize}

 
\vspace{-0.1cm}
\section{Related Work}\label{sec:related}

\textbf{Physics and deep learning.} The relationship between physics and deep learning is symbiotic: Physics inspired segmentation~\citep{lengyel2021zero} and generative models~\citep{toth2019hamiltonian, takeishi2021physics} as well as the design of new architectures~\citep{gin2021deep}. Likewise, deep learning is used to study, understand and create new physics from data~\citep{brunton2016discovering,  champion2019data, iten2020discovering, watters2017visual}.  Techniques like physics-informed neural networks (PINN)~\citep{karniadakis2021physics, RAISSI2019686} or Lagrangian neural networks (LLN)~\citep{cranmer2020lagrangian, lutter2019deep} are designed to solve inverse problems. Yet, PINNs are usually constrained to initial conditions, boundary conditions, and time reference~\citep{karniadakis2021physics, gao2022physics, meng2023pinn}. Moreover, these methods are supervised and require labeled data. Because obtaining labeled data in inverse problems is expensive or even infeasible~\citep{adam2015higgs} we avoid the need for labels by proposing an unsupervised method, following~\citep{adam2015higgs, ball2010data, libbrecht2015machine, meijering2016imagining, varoquaux2022machine}.





While some methods incorporate physics knowledge or design inspiration, they focus on future predictions rather than estimating physical parameters~\cite{watter2015embed, banijamali2018robust, rackauckas2020universal, jaques2021newtonianvae}. Alternatively, some approaches attempt to relearn or propose new equations~\cite{champion2019data, fries2022lasdi, lee2022structure}. Our method differs by using known governing equations to both predict on latent space and learn physical parameters, which prevents direct comparisons with these techniques. Moreover, some of these approaches not suitable for video applications.

\textbf{Learning physics from video.} Research on learning physics from videos often focuses on frame prediction 
~\citep{battaglia2016interaction, fragkiadaki2015learning, guen2020disentangling, kossen2019structured} and not on accurate parameter estimation. Existing works on extracting physical information from videos~\citep{NIPS2017_4c56ff4c} consider physical parameters (e.g. the friction coefficient) or the value of the dynamical state variable (e.g. the position or velocity), but employ supervised methods which require labelled datasets with access to the dynamical variables' or parameters' ground truth~\citep{NIPS2017_4c56ff4c, de2018end, lutter2019deep, RAISSI2019686, wu2015galileo, watters2017visual, zheng2018unsupervised, yang2022learning}. Moreover, these methods mainly address motion dynamics and are based on systems similar to interaction networks~\citep{ battaglia2016interaction,watters2017visual,  velivckovic2022reasoning}. Some methods~\cite {weiss2020correspondence, kairanda2022f} aim to parameterize and solve differential equations to simulate the deformations. However, dynamical systems are not limited to motion, and we propose a method that goes beyond motion, illustrated with use-cases for intensity changes and scaling.

\textbf{Unsupervised parameter estimation from video.} 
Existing unsupervised work uses frame representations and physics priors of known governing equations with unknown parameters to estimate~\citep{weiss2020correspondence, jaques2019physics, hofherr2023neural, NEURIPS2020_79f56e5e, kandukuri2020learning}. Some models~\citep{kandukuri2020learning, NEURIPS2020_79f56e5e} use variational auto-encoders (VAE)~\citep{kingma2013auto} with a physics engine between the encoder and the decoder to estimate parameters; however, the reconstructions are poor, constraining the model to simple motion problems. 
Approaches similar to ours~\citep{hofherr2023neural, jaques2019physics} estimate parameters from a single video without annotations, and constitute our baselines. Empirically, our method compares favorably in terms of robustness to initializations, analysis of latent space dynamics and extension to different systems.

\textbf{Datasets.} One dataset for unsupervised parameter estimation from video is Physics101~\cite{phys101}, which includes recordings of physical experiments, but lacks parameter ground truths and object masks required by some models~\cite{hofherr2023neural}. Consequently, Physics101 has not been used for unsupervised parameter estimation, and existing models are primarily tested on synthetic datasets, with minimal overlap across studies~\cite{jaques2019physics, hofherr2023neural, yang2022learning}. In this paper, we introduce Delfys75: a new dataset featuring real-world videos of diverse experiments, estimated physical parameters, and object masks.

\textbf{Baselines.}\label{sec:baselines} 
As a strong unsupervised parameter estimator,~\citet{jaques2019physics} uses a traditional auto-encoder with a physics engine in the latent space to reconstruct inputs and generate future frame predictions. 
The model also incorporates a U-Net in the encoder to learn segmentation masks for the object of interest in an unsupervised way; the capability to learn this mask is linked to the spatial transformer used in the decoder similar to~\citep{hsieh2018learning}. During inference, the spatial transformer is used to perform affine transformations on the mask to `move' the object in future frame predictions. This limits the application of~\citep{jaques2019physics} to motion dynamics. 

On the other hand, ~\citet{hofherr2023neural} uses a differentiable ODE solver to estimate the parameters. This model also uses a spatial transformer, but at the pixel level: Object pixels are displaced using predictions made by the ODE solver. The model needs to be trained with the use of masks, to learn which pixels are displaced in future frames.

Taken together, reconstructing frames using only low dimensional data (e.g. a set of positions and velocities) is challenging and slows down training for parameter estimation. Therefore,~\citep{hofherr2023neural, jaques2019physics, weiss2020correspondence} had to limit their scope to using a mask and a spatial transformer~\citep{hsieh2018learning}, excluding dynamical systems with changes in intensity and colour, deformations and non-uniform scaling among others which we explicitly allow in our paper.

\vspace{-0.1cm}
\section{Model}\label{sec:methods}

\begin{figure}[t]
    \centering
    \includegraphics[width=0.5\textwidth]{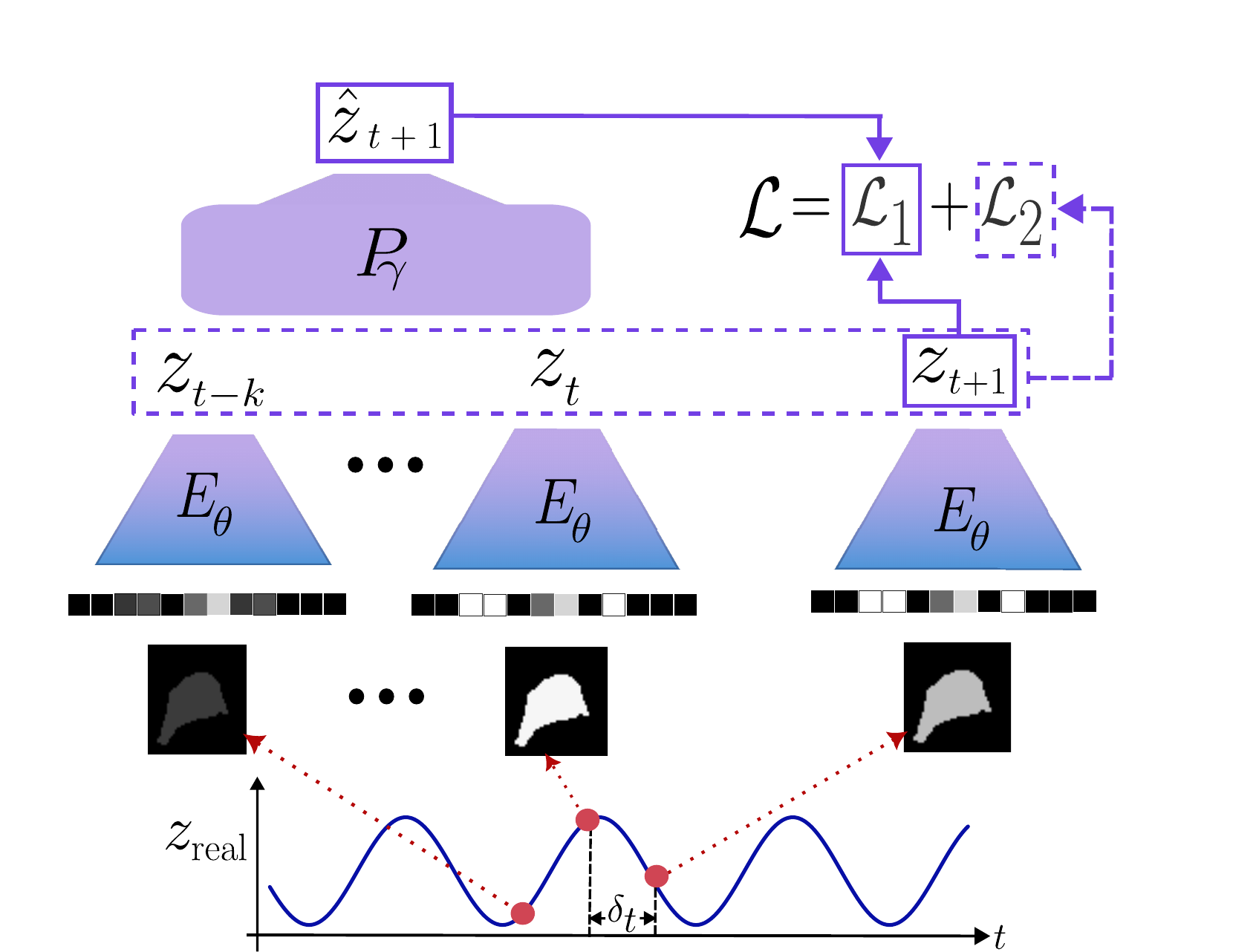}
    \caption{Method overview. A video recording of an object with a periodic brightness change (bottom) displays dynamics $z_{\text{real}}$ with sampling period $\delta t$.  Each frame is mapped by the encoder $E_\theta$ to the unsupervised latent representation $z_t$. The physics block $P_\gamma$ generates a prediction of the future step $\hat{z}_{t+1}$, which we compare to the encoded representation $z_{t+1}$ of frame $t{+}1$. \textbf{Top-right:} Loss function of our model;  the first term ensures the prediction fits with the encoding, while the second expression controls the variance of $z$. This image summarizes our methodology and the relationship between the different blocks.}
    \label{fig:arqu}
    \vspace{-0.2cm}
\end{figure}

Our model estimates the parameters of a known governing equation from a video recording. Since it is unsupervised, the training set consists of unannotated frames with a known frame rate denoted as $\delta t$. We do not reconstruct frames and thus have only a simple encoder and a physics block.   
In Figure~\ref{fig:arqu} we show our approach. 


\textbf{Scope.} We study systems represented by autonomous differential equations (Eq.~\ref{eq:autonODE}), which depend only on the state variable captured in video. Assuming no external forces,
\begin{equation}
    z^{(n)}+ \gamma_{n} z^{(n-1)} + ... + \gamma_{2} z^{(1)} + \gamma_{1} z + \gamma_{0} = 0 |  z^{(k)} = \frac{d^{k}z}{dt^{k}}
   \label{eq:autonODE}
\end{equation}
is an $n^\textrm{th}$-order system, where $z$ is the time-dependent state variable or "dynamic variable", $z^{(k)}$ with $k = 1, 2,\dots n$  is the $k^\textrm{th}$-derivative of $z$ with respect to time $t$ and $\gamma_i$ with $i = 0,1,\dots n-1$ are the parameters of the equation we want to estimate. 

While our approach can be extended to equations of arbitrary order, in the following \emph{proof-of-concept}, we first consider a second-order differential equation since it is the maximum order used in previous work:
\begin{equation}
   z^{(2)}+ \gamma_1 z^{(1)} + \gamma_0 z = 0. 
   \label{eq:secondOrder}
\end{equation}

\noindent \textbf{Physics block.} Our physics block numerically solves the differential equation using a single step of Euler's method:
\begin{eqnarray}
\label{eq:eulers}
    &z^{(1)}_t \approx \frac{ z_{t+1} - z_{t} }{ \delta t  } \approx \frac{ z_{t} - z_{t-1} }{ \delta t  }  \\ [4pt]
    &z_{t+1} = z_{t} + \delta t   z^{(1)}_{t}   \\ [5pt]
    &z^{(1)}_{t+1} = z^{(1)}_{t} + \delta t z^{(2)}_{t}.
\end{eqnarray}
Plugging in Eq.~\ref{eq:secondOrder}, we can rewrite the physics block as:
\begin{equation}
   \hat{z}_{t+1} = z_{t} + \delta t \left( z^{(1)}_{t} - \delta t ( \gamma_1 z^{(1)}_t + \gamma_0 z_t ) \right).
   \label{eq:phys2or}
\end{equation}
\begin{equation*}
  P : \mathbb{R}^d \rightarrow \mathbb{R}^d,
\end{equation*}
\begin{equation}
   \hat{z}_{t+1} = P_\gamma(z_{t}, .. z_{t-n}; \bf{\gamma} ).
   \label{eq:physblock}
\end{equation}
where $\gamma_i$ are learnable parameters and the predicted latent space $\hat{z}$ for time $t+1$ is a function $P_\gamma(\cdot  )$ of the latent representations of the $n$ previous frames. We use the notation $ \mathbf{\hat{z}}$ for the latent space predicted using the function $P$, which is different from $\mathbf{z}$ predicted by the encoder. 

\noindent \textbf{Encoder.} The encoder is a network $E_\theta(x)$ that maps images $x \in \mathbb{R}^{w\times h \times  c}$ to the state variable $z \in \mathbb{R}^d$, where $d$ is the number of dynamic variables. We use an MLP, similar to~\cite{jaques2019physics} suitable as localization networks, with three layers and the ReLU activation function for the mapping
\begin{eqnarray}
  z_t  = E_\theta(x_t), \quad  \text{where} &E : \mathbb{R}^{w\times h \times  c} \rightarrow \mathbb{R}^d. 
   \label{eq:E}
\end{eqnarray}

\textbf{Predictions.} The encoder maps images $x_t \in \mathbb{R}^{w\times h \times  c}$ to the state variable $z_t \in \mathbb{R}^d$ 
for all time steps $t \in [0,T]$ leading to $\hat{z}$ with dimensionality 
$\mathbb{R}^{T\times d }$ for all input frames. In particular, we have the following two vectors:
    \begin{equation}
\mathbf{z}=
       \begin{bmatrix}
        z_{n}\\
        \vdots\\
        z_{t+1}\\
        \vdots          \\
        z_{T}
        \end{bmatrix} 
        = 
        \begin{bmatrix}
        E_\theta(x_n )\\
        \vdots\\
        E_\theta(x_{t+1} )\\
        \vdots          \\
        E_\theta(z_{T} )\\
        \end{bmatrix}   \\
    \end{equation}
\begin{equation}
\mathbf{\hat{z}}=
    \begin{bmatrix}
        \hat{z}_{n}\\
        \vdots\\
        \hat{z}_{t+1}\\
        \vdots          \\
        \hat{z}_{T}
        \end{bmatrix} 
        = 
        \begin{bmatrix}
        P_\gamma(z_{n-1}, \dots, z_{0}; \mathbf{\gamma}  )\\
        \vdots\\
        P_\gamma(z_{t}, \dots,z_{t-n}; \mathbf{\gamma} )\\
        \vdots          \\
        P_\gamma(z_{T-1}, \dots,  z_{T-n}; \mathbf{\gamma} )\\
        \end{bmatrix}   \\
    \end{equation}



\textbf{Loss function.}\label{sec:LossFunction} 
The first goal of the loss function is to minimize the difference between the predictions $\mathbf{z}$ and $\mathbf{\hat{z}}$ over a batch of size $M$, as $\mathcal{L}_1 = \frac{1}{M} \sum_{i=1}^{M} (z_i - \hat{z}_i)^2$, \ie,
\begin{equation}
    \frac{1}{M} \sum_{i=1}^{M} \Bigl[ E_\theta(x_i) - P_\gamma \bigl(E_\theta(x_{i-1}), \dots,E_\theta(x_{i-n})  \bigr) \Bigr] ^2.
    \label{eq:L1}
\end{equation}
One problem with this approach is the convergence to the trivial solution such that ${E_\theta(x) = 0 \; \forall x}$ and ${P_\gamma(z) = 0  \; \forall z}$. 
To avoid this problem, we induce variance in the encoder's output. Inspired by  the VAE~\citep{kingma2013auto} we encourage ${z_j \in \mathcal{N}(\mu, \sigma^2)}$. The values of $\mu$ and $\sigma^2$ effectively define the range of $z$ by renormalization of the metric. 
As in the VAE~\citep{kingma2013auto}, we define the second part of the loss function using the Kullback-Leibler divergence (KL-divergence). Thus, $z_j$ is a sample of the random variable $Z\sim \mathcal{N}(\mu_z, \sigma_z^2)$ and we want it to follow a particular prior distribution $Q\sim \mathcal{N}(0, 1)$. Then KL-divergence is given by:
\begin{equation}
    \mathcal{L}_2 = \text{KL}(Z||Q) = - \frac{1}{d} \sum_{j=1}^{d} \bigl(1 + \ln( \sigma_{z}^2 ) - \mu_{z}^2 -  \sigma_{z}^2 \bigr).
    \label{eq:L2}
\end{equation} 
Here, the KL-divergence is used differently than conventional: VAEs~\citep{kingma2013auto} typically assume the encoder finds the correct distribution, and use the sampling trick to obtain the decoder input. In our proposal, we do not sample from the latent distribution. Instead, we constrain the encoder to learn the dynamical state variable. Thus, we calculate the mean $\mu_{z}$ and variance $\sigma_{z}^2$ over the batch in the loss.

Finally, combining the two loss functions:
\begin{multline}
    \mathcal{L} = \mathcal{L}_1 + \mathcal{L}_2 =  \frac{1}{M} \sum_{i=1}^{M} (z_i - \hat{z}_i)^2 \\ - \frac{1}{d} \sum_{j=1}^{d} \bigl(1 + \ln( \sigma_{z}^2 ) - \mu_{z}^2 -  \sigma_{z}^2 \bigr).
    \label{eq:full_loss}
\end{multline}
Our loss function (Eq.~\ref{eq:full_loss}) serves several purposes: First, $E_\theta(\cdot )$ maps images to random variables $z$ in latent space, and $L_2$ encourages that $z$ is normally distributed. Then, the function  $P_\gamma(\cdot )$ preserves the sense of order and relation between latent spaces from different frames. Since we are analyzing a single video, $z$ contains information about the temporal properties of the sequence of frames. Finally, $L_1$ makes the model consistent since the physical predictions made by the physics block  $P_\gamma(\cdot )$ are generated by $E_\theta(\cdot)$.

\textbf{Training.} For parameter estimation in the physics block   $P_\gamma(\cdot )$, we used a learning rate proportional to the initial value $\gamma_i^0$ of the learnable parameter $\gamma_i$, where $\text{lr}_{(\gamma_i)} \sim 10^{[\log_{10} |(\gamma_i^0)| ]} $. This approach provides sufficiently large step sizes at the beginning of training to escape local minima. More details  about the optimizer and hardware are in the supplementary material.

\vspace{-0.1cm}
\section{Delfys75: a new real-world dataset}\label{sec:dataset}

\begin{figure}[h]
    \centering
    \includegraphics[width=\linewidth]{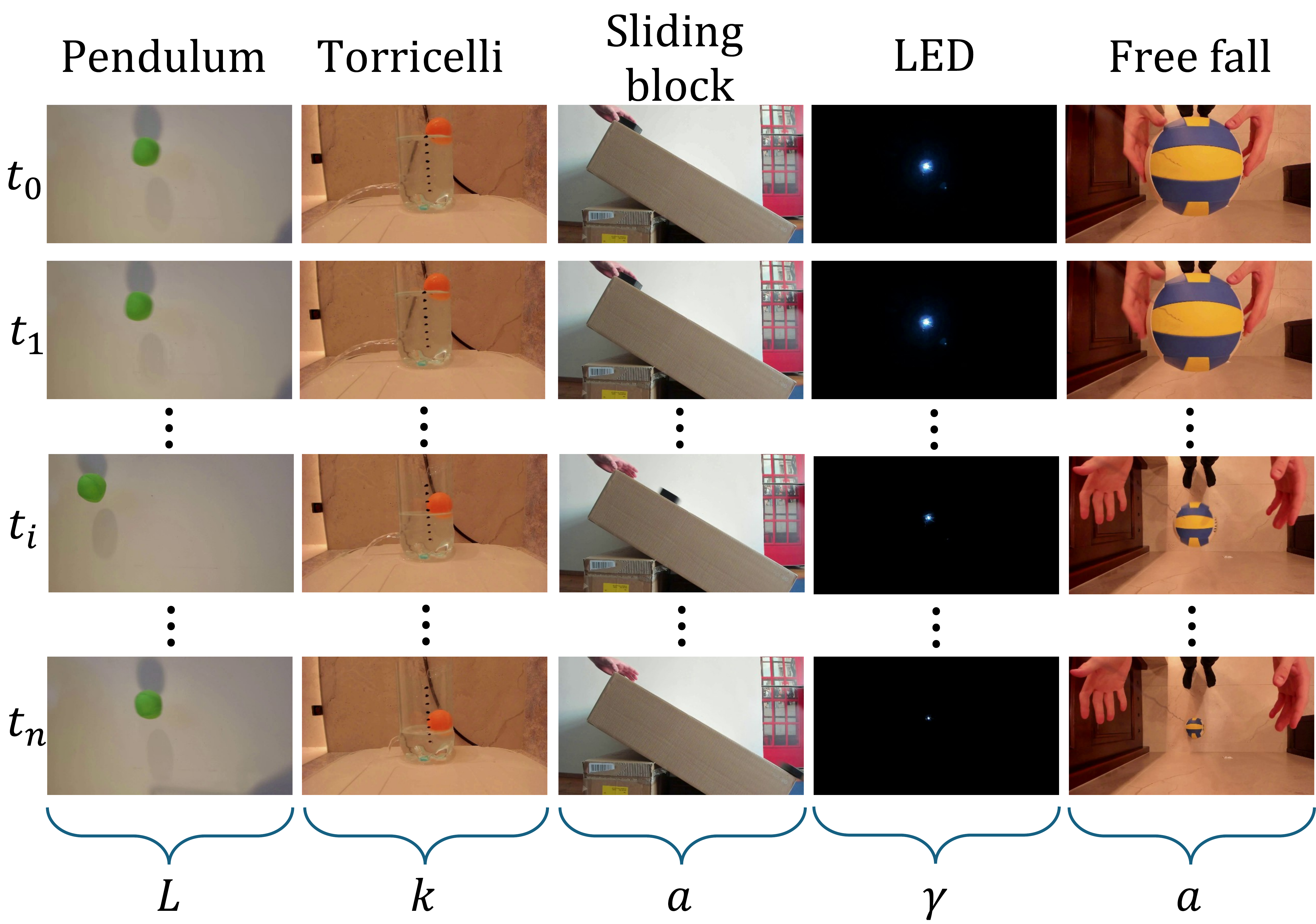}
    \caption{ Delfys75 is the first, real-world physical parameter estimation dataset. Top to bottom: first ($t_0$), second ($t_1$), middle ($t_i$), and last frames ($t_n$). Specified at the bottom are the estimated parameters in each scenario. Note the complex shadows, shading, and realistic lighting conditions, in a natural environment.}
    \label{fig:dataset_visualization}
\end{figure}

Delfys75 contains five experiments: 3 with motion, one with brightness and one with scale dynamics.  Motion is overrepresented to enable a deeper comparison with existing methods, which cannot be evaluated on brightness or scale dynamics. Each experiment is recorded in 3 different settings, with five videos per setting, resulting in a dataset of 75 videos. The resolution is $1920 \times 1080$ at 60 frames per second. Sample frames are visualized in Figure~\ref{fig:dataset_visualization}.

\textbf{Object masks} are binary segmentation masks of the object undergoing the physical transformation. Masks can be learned~\cite{jaques2019physics} or otherwise need to be provided~\cite{hofherr2023neural} with the training data. In our dataset, we provide object masks per frame, annotated using SegmentAnything2~\cite{ravi2024sam},  since some baseline methods require it~\cite{hofherr2023neural}. Our method does not need object masks.

\textbf{Parameter ground truths} and estimation errors, are obtained via manual physical parameter estimation. See the supplementary material for details.

 
\subsection{Scenarios}
\vspace{-0.1cm}
The \textbf{pendulum} is a classical motion-based system~\cite{hofherr2023neural, jaques2019physics, yang2022learning} where the position of the pendulum is expressed by the angle $\theta$ and the dynamics are given by:
\vspace{-0.1cm}
\begin{equation}
    \theta^{(2)} + \zeta \theta^{(1)} + \frac{g}{L} \sin (\theta) = 0.
\vspace{-0.1cm}
\end{equation}
with $\zeta$ being the damping factor, $g$ the gravitational acceleration and $L$ the length of the pendulum. We vary the length of the pendulum to generate the different settings.

The \textbf{Torricelli} motion describes the change in water height over time in a container with an orifice through which water is draining. In this setup, a floating ball at the surface serves as an indicator of the water level. Our objective is to determine the constant $k$, related to the water flow rate, from the video as the water level decreases

\vspace{-0.1cm}
\begin{equation}
    h^{(1)} = k \sqrt{h}.
\vspace{-0.1cm}
\label{eq:torricelli}
\end{equation}

The \textbf{sliding block} involves the movement of an object sliding down a ramp~\cite{phys101, hofherr2023neural}, with dynamics given by:
\vspace{-0.1cm}
\begin{equation}
    x^{(2)} = g(\sin(\alpha) - \mu \cos(\alpha)).
\vspace{-0.1cm}
\end{equation}
where $x$ is the position with $\alpha$ the inclination angle of the ramp, $\mu$ the friction coefficient and $g$ is the gravity. Each of our three settings employs a different inclination angle.


In the \textbf{LED} experiment, the brightness of an LED lamp is varied. The intensity $I(t)$ of the LED is adjusted following $I(t) = e^{-\gamma t}$ with a controllable decay $\gamma$. We disable camera auto-focus, auto-brightness, and white balancing for this experiment to ensure consistent brightness. $\gamma$ is changed between the three settings.

In the \textbf{free fall} scenario, a ball dropped from a height $h_0$ is filmed from above, causing the ball's apparent radius $r(t)$ to decrease as it falls. Using similar triangles, $r(t)$ is described by:
\vspace{-0.2cm}
\begin{equation}
    r(t) = \frac{r_0 f}{h_0 + \frac{gt^2}{2}}.
\vspace{-0.1cm}
\end{equation}
with $r_0$ the real-life radius of the ball and $f$ the focal length of the camera. We disable camera auto-focus to ensure constant $f$.
The radius $r_0$ is modified between settings.

The videos contain hand interactions, which introduce occlusions and variability. The pendulum includes border occlusions and cast shadows. Torricelli has background changes due to water flow, ball rotation, and deformation. The free fall presents perspective distortions. In a a similar way, LED suffers from reflections and non-uniform intensity changes. These factors add complexity and affect the performance of the model. 

\vspace{-0.2cm}
\section{Experiments}
     

\begin{figure*}[ht]
         \centering
         \begin{subfigure}[b]{0.60\textwidth}
         \centering
         \includegraphics[width=\textwidth,trim=0 20 0 20]{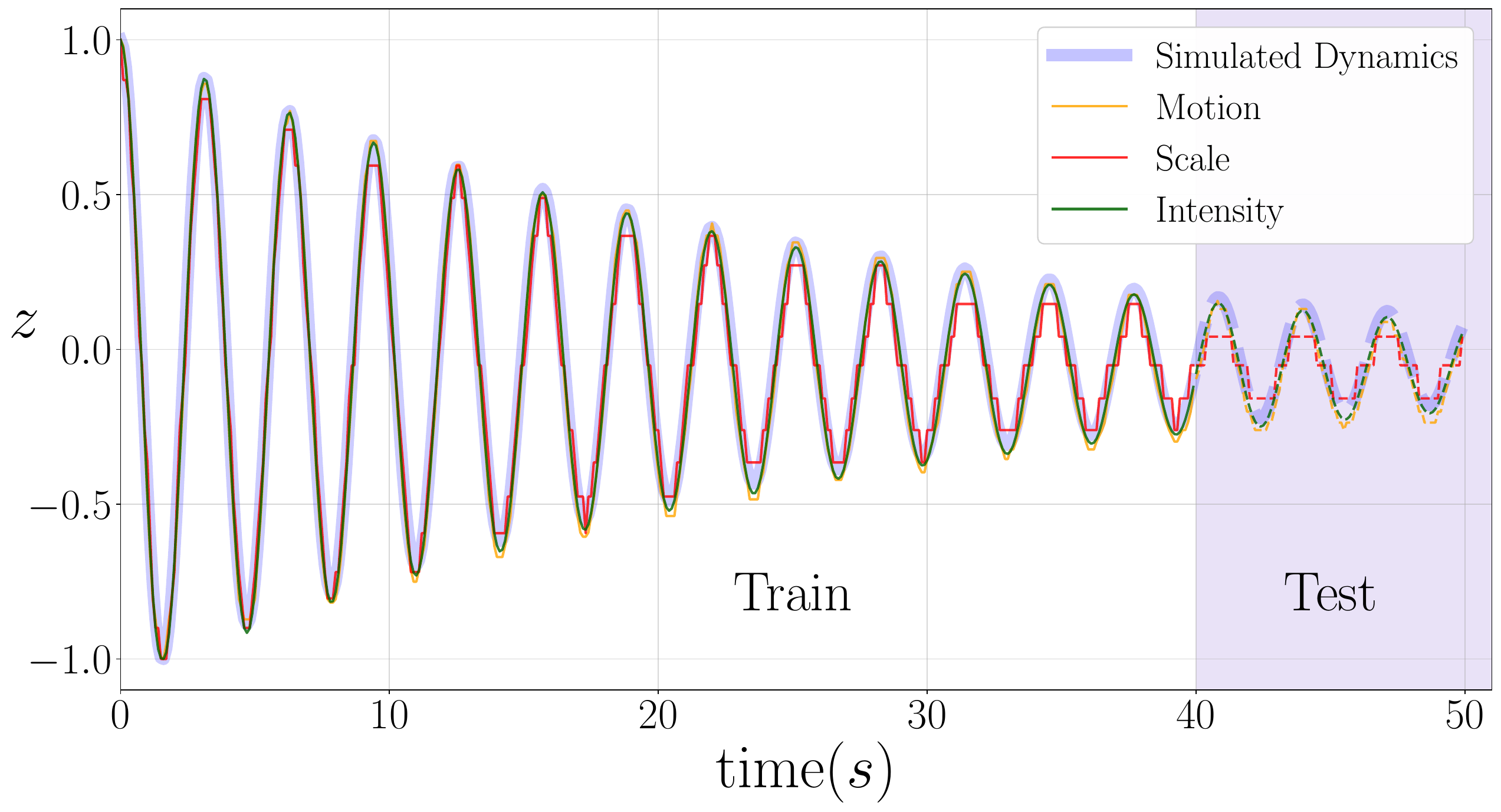}
         \caption{}
         \label{fig:LatenSpacePred_1}
     \end{subfigure}
     \hfill     
     \begin{subfigure}[b]{0.38\textwidth}
     
         \centering
            \begin{tabular}{ccc}            
            \toprule
            \multirow{2}{*}{\textbf{ Equation}}&\multicolumn{2}{c}{$z^{(2)} +\gamma_1 z^{(1)} + \gamma_0 z= 0$}\\
            \cmidrule{2-3}
            & \multicolumn{1}{c|}{$\gamma_0$} & \multicolumn{1}{c}{$\gamma_1$}\\
            \midrule
            \multicolumn{1}{l|}{Motion} & \multicolumn{1}{c|}{$3.943 \pm 0.008$} &\multicolumn{1}{c}{$0.144 \pm 0.007$}\\
            \multicolumn{1}{l|}{Intensity} & \multicolumn{1}{c|}{$3.887 \pm 0.035$} &\multicolumn{1}{c}{$0.089 \pm 0.010$}\\
            \multicolumn{1}{l|}{Scale} & \multicolumn{1}{c|}{$ 4.055 \pm 0.026$} &\multicolumn{1}{c}{$0.910 \pm 0.012$}\\
            \multicolumn{1}{l|}{\textbf{GT}} & \multicolumn{1}{c|}{$4.0016$} &\multicolumn{1}{c}{$0.08$}\\
            \bottomrule
            \end{tabular}
        \vspace{1.5cm}            
         \caption{}
         \label{tb:ExperimentsBase}         
     \end{subfigure}     
         \caption{\textbf{(a)} Latent space estimation of the dynamic variable $z$ for the three synthetic datasets. The blue line shows the `ground truth' value $z_{\text{real}}$ of the simulated dynamics. The model was trained with the dynamics of the continuous line while the dashed lines show the ground truth (blue) and predictions (yellow, red, green) on the extrapolated test set. \textbf{(b)} Parameter estimation accuracy. Rows 1-3: mean $\pm$ standard deviation of each learnable parameter in the physics block after training, bottom row: ground truth (GT). The values are obtained over 7 different runs with different initializations. We observe good agreement between the predicted and ground truth dynamics.
         }
         \label{fig:LatenSpacePred}
         \vspace{-0.4cm}
\end{figure*}

\subsection{Synthetic Video Datasets}\label{sec:fullyControlExp}

We start in a fully-controlled setting with three synthetic datasets involving motion, intensity, and scale, visualized in Figure~\ref{fig:baseexp}. We describe dataset details in the supplementary material. 
State-of-the-art methods for physical parameter estimation commonly use simulated datasets~\citep{de2018end, jaques2019physics, velivckovic2022reasoning, watters2017visual, yang2022learning, zheng2018unsupervised}, where objects appear in different colors on a black background. The pendulum system is a typical case of study, but for scale and intensity, there were no baselines available.

\textbf{Motion.} We used a pendulum model popular in literature~\cite{jaques2019physics, hofherr2023neural, brunton2016discovering}. In this dataset, the state variable is the angle of the pendulum $z_{real}=\theta$ $ \in [-180.0^\circ,180.0^\circ]$.

\textbf{Intensity.} We consider time-varying grayscale pixel intensity of an irregular shape. We normalized the intensity dynamics to be in the range $z_{\text{real}} \in [0.2,1.0] $.

\textbf{Scale.} We use a filled circle centered in the middle of the image, where the radius is proportional to the dynamic variable. However, the scaling transformation is not symmetric; while one half of the circle grows, the other half becomes smaller and vice versa. We normalized the range of the radius dynamics between $z_{\text{real}} = r \; \in[-10,10]$.

\begin{figure}[H]
    \centering
    \includegraphics[width=0.48\textwidth]{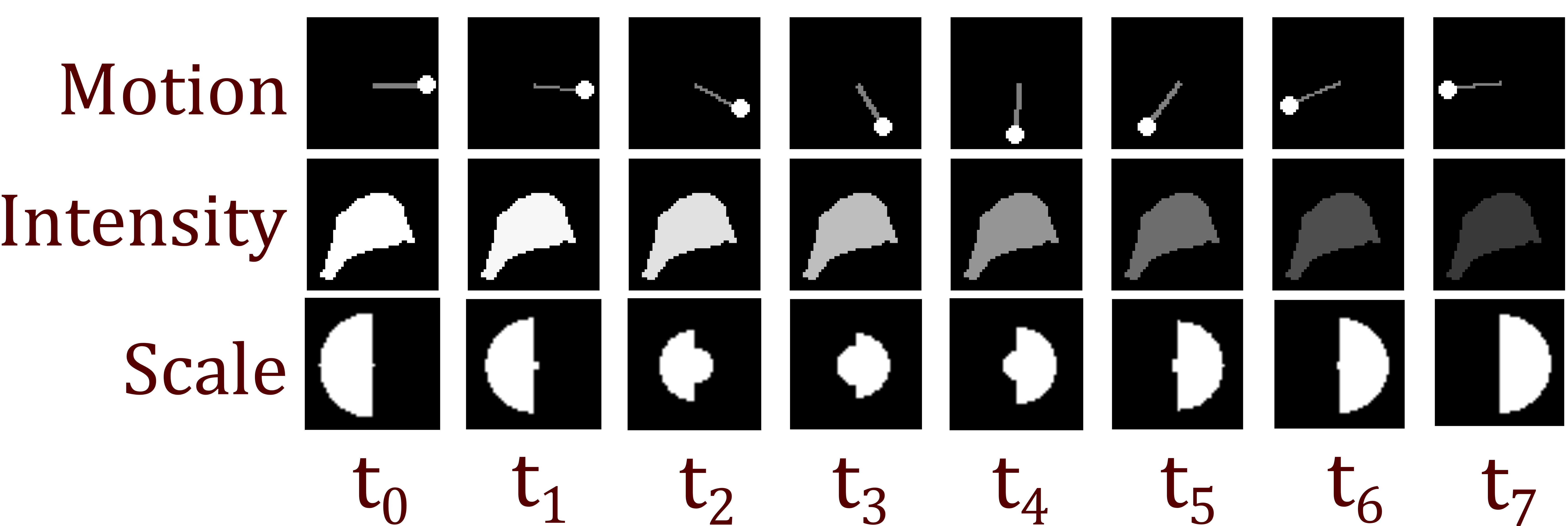}
    \caption{Example frames from the synthetic datasets. Each row shows a different dataset, corresponding to a different continuous dynamical system, and each column a different time sample.}
    \label{fig:baseexp}
    \vspace{-0.5cm}
\end{figure}

\paragraph{Dynamics Representation in Latent Space}
\label{sec:latent}
First, we train our model using the three synthetic datasets and evaluate the dynamics $z$ estimated by the encoder $E_{\theta}(\cdot)$. We compare $z$ to its ground truth value $z_{real}$, which was used to generate the data. Following Eq.~\ref{eq:secondOrder}, we consider second-order dynamics for all datasets, where the evolution of the state variable $z$ follows a dampened oscillation.


Figure~\ref{fig:LatenSpacePred} shows our model is capable of estimating the dynamics $z$ for all three datasets. Although understanding oscillations in data can be challenging for neural networks~\cite{xu2019frequency}, our unsupervised loss fits the dynamical behaviour reasonably, with small deviations from the ground truth. 

Importantly, our model has complete physical interpretability since we use the differentiable, second-order ODE (Eq.~\ref{eq:secondOrder}) in the physics block. Due to this physics prior, the model is able to generalize at test time to previously unseen time steps. The network's extrapolation of $z$ to unseen future time steps is shown using dashed lines in Figure~\ref{fig:LatenSpacePred_1}.

We observe that the most accurate results are obtained using the `Intensity' dataset. 
We attribute this to discretization error: 8-bit inputs of pixel intensity can assume $256$ different values. In contrast, with motion and scale, the dynamics are discretized by the pixel locations. In particular, for a $(50 \times 50)$ frame size, the discretization of the dynamic variable is increasingly more impactful as the oscillation amplitude is decreased. This effect is seen clearly in the latent space dynamics of the `Scale' dataset in Figure~\ref{fig:LatenSpacePred}a (red line), which displays discrete jumps. The discretization error also disproportionally affects the parameter estimation in the `Scale' dataset, specifically the prediction of the parameter $\gamma_{1}$, which is discussed in the next section.

\vspace{-0.4cm}
\paragraph{Parameter Estimation Accuracy}
\label{sec:paramAcc}

To evaluate our model's accuracy in estimating parameters $\gamma$, we consider second-order equations defined in Eq.~\ref{eq:secondOrder}, where $\gamma_0$ is the frequency and $\gamma_1$ is the damping factor of the oscillations. (See supplement for details.)

Figure~\ref{fig:LatenSpacePred}b shows the $\gamma$ values learned by our model. We find that the model can accurately estimate $\gamma_0$ for all systems. Figure~\ref{fig:LatenSpacePred}a serves as validation that the model indeed fits the correct frequency. The varying accuracy for $\gamma_1$ is attributed to the discretization error discussed in Section~\ref{sec:latent}, as the accuracy decrease correlates well with the increased discretization in the motion and scale datasets.


\vspace{-0.1cm}
\subsection{Robustness and Stability}
While previous work can reliably generate frame reconstructions using physics blocks in latent space, they often lack an analysis of the parameter estimation. 
Thus, it is not known if the generated frames correctly use the latent space information in an interpretable manner. In fact, it is known that the baseline models are sensitive to initialization and may fail to converge~\citep{hofherr2023neural, Jaques2019git}. In this section, we evaluate the robustness of our model against changes in parameter initializations. 

\begin{figure}[h]
\centering
\includegraphics[width=0.37\textwidth,trim=0 20 0 10]{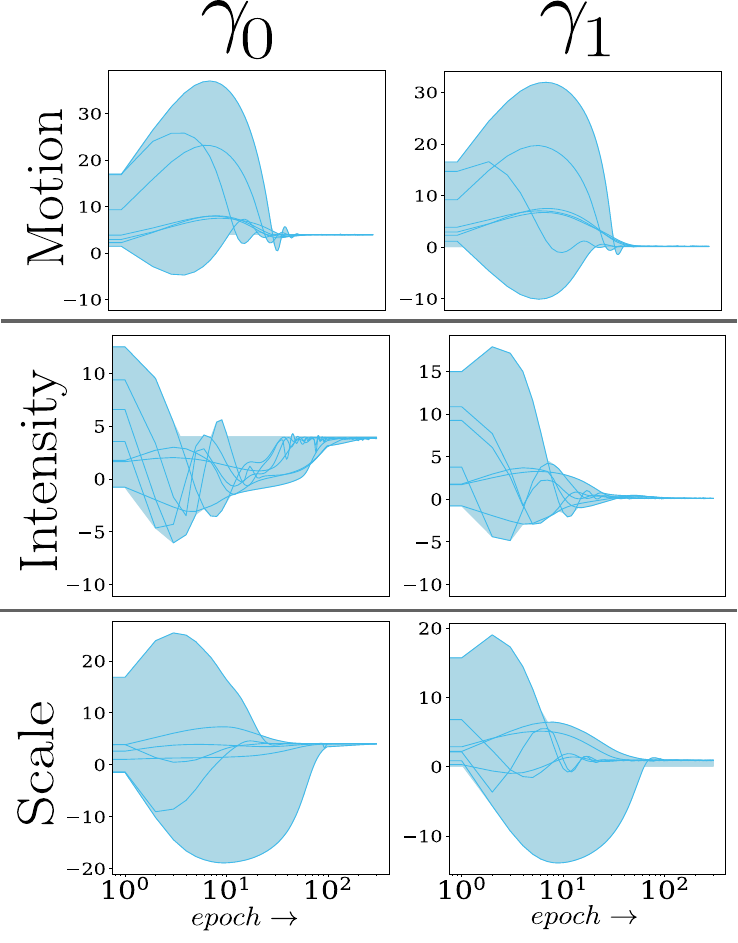}
\caption{Robustness of the parameter estimation against different initializations. The rows indicate the different dynamical systems, while the columns are the parameters to estimate. The vertical axis corresponds to the parameter value, and the horizontal axis is the epoch. Blue lines show the value of the estimated parameter $\gamma_i$ over training epochs. Since convergence was relatively fast, the horizontal axis is on a logarithmic scale for visibility. The shading highlights the variance of the trajectories before convergence.} 
\label{fig:Robustness}
\vspace{-0.3cm}
\end{figure}

We initialize the learnable parameters $\gamma$ in the interval $[-10.0, 10.0]$ over 7 runs. Ground truth values were ${\gamma_0 = 4}$ and ${\gamma_1 = 0.08}$. Figure~\ref{fig:Robustness} shows the convergence of the parameter estimation during training with different initializations. While the trajectories may initially diverge, the trained model consistently converges to the ground truth values for each dataset, independent of the initialization. The relatively low standard deviations of the final $\gamma$ estimates in Figure~\ref{fig:LatenSpacePred}b highlights the stability of our model.

\begin{table}[h]
\centering
\begin{tabular}{m{1.5cm}>{\centering\arraybackslash}m{1.0cm}>{\centering\arraybackslash}m{1.3cm}>{\centering\arraybackslash}m{1.3cm}c}
\toprule
\multicolumn{1}{c}{} & PAIG \cite{jaques2019physics} & PAIG w/o U-Net & NIRPI \cite{hofherr2023neural} & Ours \\
\cmidrule{2-5}
Parameters &5.27M  &4.78M &75.42K &4.19M \\
\midrule
Time per $\,\,$ Epoch $[s]$ &252.72&80.56&0.11&0.95 \\
\midrule
Decoder&\Yesred &\Yesred &\Yesred &\Nogreen  \\
\midrule
Mask&\Nogreen  &\Yesred &\Yesred &\Yesred \Nogreen  \\
\bottomrule
\end{tabular}
\caption{Comparison of baseline models evaluated on the public dataset from~\citep{jaques2019physics}. Our model only needs a mask if there are multiple objects present in the video.}
\label{tb:baselineComp}
\vspace{-0.3cm}
\end{table}

\vspace{-0.1cm}
\subsection{Baseline Comparison on Synthetic Data}
\label{sec:baselinesComparision}
The baselines PAIG~\citep{jaques2019physics} and NIRPI~\citep{hofherr2023neural} are not designed to handle the intensity and scale settings of our synthetic datasets. Therefore, for a fair comparison, we use the synthetic dataset first proposed in~\citep{jaques2019physics} and reused in~\citep{hofherr2023neural}. It consists of two MNIST digits moving following the spring equation parametrized by the spring constant $k$; the dataset was created using $k=2$. Details of the governing equation and visualizations are given in the supplement. In this dataset the dynamics is given by the $(x,y)$ position of each digit, therefore the dimensionality of the latent space is $d =4$.

Table~\ref{tb:baselineComp} presents a size comparison of the models along with training time per epoch. PAIG~\cite{jaques2019physics} employs a U-Net to learn the object masks and does not require them as input. Other baselines, as well as our model for multiple objects, need masked inputs and, therefore, do not employ a segmentation block. We also consider PAIG~\cite{jaques2019physics} with masked input and without the U-Net block in our comparisons.

\begin{figure}[h] 
    \centering
    \includegraphics[width=0.45\textwidth,trim=0 30 0 30]{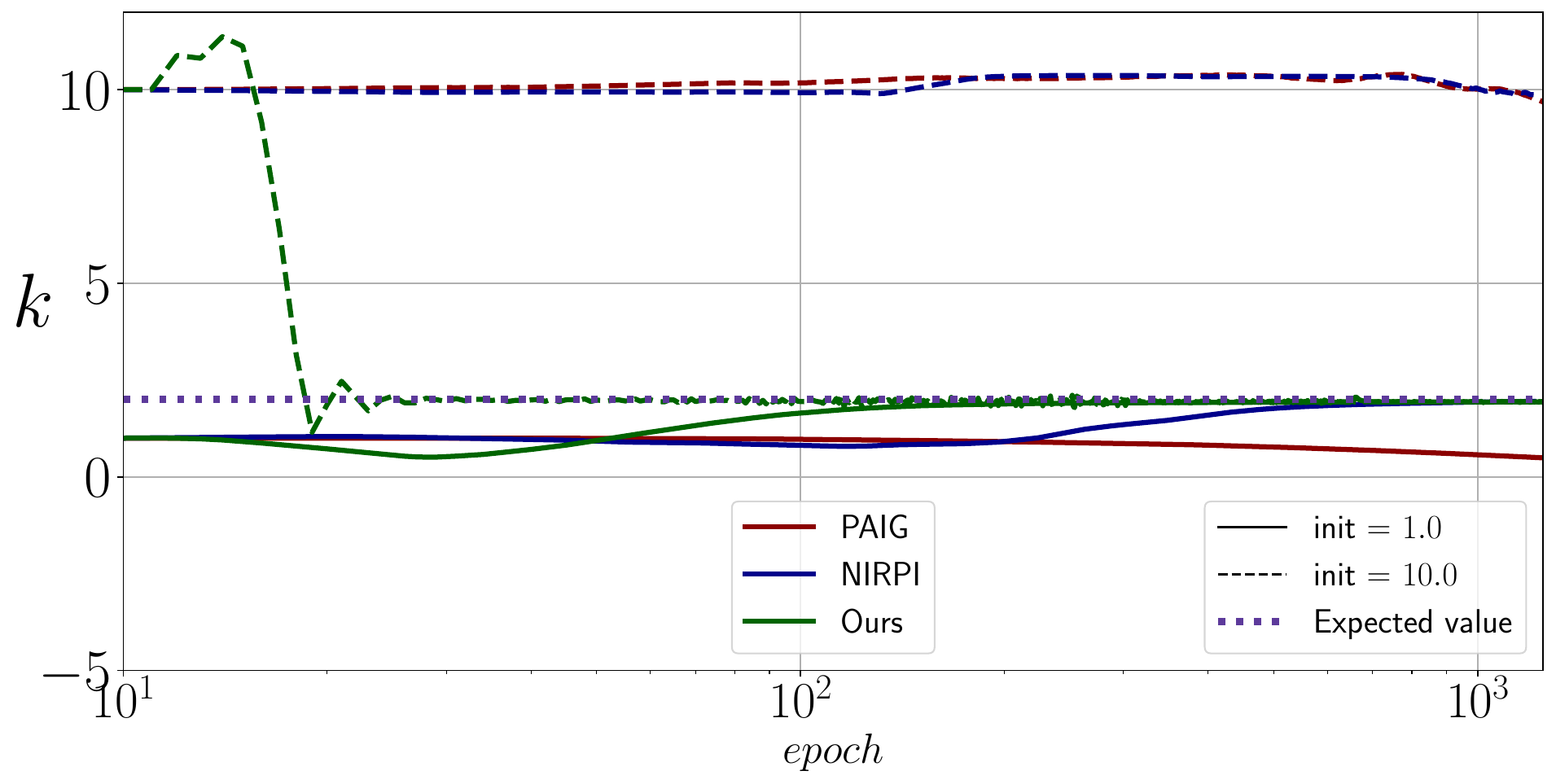}
    \caption{Robustness of the parameter estimation compared to baselines. For a fair comparison, we use the synthetic dataset created originally by the authors of the baseline papers to evaluate their models~\citep{jaques2019physics}. We plot the trajectories of the estimated parameter $k$ during training with different initializations for our model (green) and for the two baseline models (red, blue). Dotted lines correspond to an initial value of $k = 10.0$, and solid lines to $k = 1.0$. Our model converges robustly to the ground truth value of $k=2.0$.}
    \label{fig:k_comparison}
    \vspace{-0.3cm}
\end{figure}

We empirically demonstrate the models' sensitivity to initial conditions, we consider PAIG~\cite{jaques2019physics} as proposed by the authors with the U-NET. We train each model twice and initialize the estimated parameter $k$ with values $1.0$ and $10.0$, respectively. The expected value is $k = 2$~\citep{jaques2019physics, hofherr2023neural}. In Figure~\ref{fig:k_comparison}, we observe that different initializations fail to converge to the correct value of $k$ for the baselines, while our model is consistent and converges to the desired value accurately. This experiment demonstrates that our model can also successfully tackle problems with multiple objects.

\begin{table*}[t]
\begin{subtable}[t]{\textwidth}
    \centering
    \scriptsize
    \begin{tabular}{c|ccc|ccc|ccc}
        \toprule

        \multicolumn{1}{c}{} & \multicolumn{3}{c}{Pendulum} & \multicolumn{3}{c}{Torricelli} & \multicolumn{3}{c}{Sliding block}  \\
        
        \multicolumn{1}{c}{} & \multicolumn{3}{c}{$L \left[m\right]$} & \multicolumn{3}{c}{$k \left[\frac{\sqrt{m}}{s^2}\right]$} & \multicolumn{3}{c}{$a \left[\frac{m}{s^2}\right]$}  \\
        
        \cmidrule(lr){2-4} \cmidrule(lr){5-7} \cmidrule(lr){8-10}  
         
        PAIG &$1.01 \pm 0.03$  &\cellcolor{moderategrey}$1.01 \pm 0.04$ &$1.01 \pm 0.04$ &$0.99 \pm 0.01$ &$0.99 \pm 0.01$ &$0.97 \pm 0.02$ &$0.35 \pm 0.03$ &$0.38 \pm 0.02$ &$0.37 \pm 0.04$  \\
        
        NIRPI &$0.77 \pm 0.33$ &$0.84 \pm 0.53$ &$0.63 \pm 0.38$ &$0.21 \pm 0.03$ &$0.14 \pm 0.04$ &$0.16 \pm 0.01$ &$-0.09 \pm 0.88$ &$-0.01 \pm 0.5$ &$-0.06 \pm 0.01$   \\
        
        Ours &\cellcolor{moderategrey}$0.51 \pm 0.01$ &$1.07 \pm 0.2$ &\cellcolor{moderategrey}$1.30 \pm 0.02$ &\cellcolor{moderategrey}$0.0094 \pm 4e^{-4}$ &\cellcolor{moderategrey}$0.0132
 \pm 5e^{-4}$ &\cellcolor{moderategrey}$0.0167 \pm 4e^{-4}$ &\cellcolor{moderategrey}$1.29 \pm 0.1$ &\cellcolor{moderategrey}$2.70 \pm 0.09$ &\cellcolor{moderategrey}$3.44 \pm 0.19$  \\
        
        \textbf{GT} & $0.45$ & $0.90$ &$1.50$ &$0.0095 $ &$0.0128 $ &$0.0162 $ & $1.441 $ & $2.300$ & $3.141$  \\
       
        \bottomrule
    \end{tabular}
    \caption{}
    \label{tb:realmotion}
\end{subtable}
\begin{subtable}[t]{\textwidth}
    \centering
    \footnotesize
    \begin{tabular}{c|ccc|ccc}
        \toprule

        \multicolumn{1}{c}{}& \multicolumn{3}{c}{LED} & \multicolumn{3}{c}{Free fall scale}\\
        
        \multicolumn{1}{c}{}& \multicolumn{3}{c}{$\gamma$}& \multicolumn{3}{c}{$a \left[\frac{m}{s^2}\right]$} \\

        \multicolumn{1}{c}{}&&&\multicolumn{1}{c}{}&
         Small ball&
         Medium ball &
         Large ball\\  
        
        \cmidrule(lr){2-4} \cmidrule(lr){5-7}

         \multicolumn{1}{c|}{Ours}&
         $2.24 \pm 0.36$&
         \multicolumn{1}{c}{$0.97\pm 0.04$}&
         \multicolumn{1}{c|}{$0.41\pm0.04$}&
         $15.0 \pm 2.1$ &
         \multicolumn{1}{c}{$9.51 \pm 1.27$} &
         $10.22 \pm 1.21$\\        

         \multicolumn{1}{c|}{\textbf{GT}}&
         $2.3$ &
         \multicolumn{1}{c}{$0.92$} &
         \multicolumn{1}{c|}{$0.46$}&$9.8$ &
         \multicolumn{1}{c}{$9.8$}& 
         $9.8$\\
         
        \bottomrule
    \end{tabular}
    \caption{}
    \label{tb:realours}
\end{subtable}    
    \caption{Parameter estimation accuracy on real-world videos. The table shows how closely each model estimates the ground truth (GT) parameter values, with mean and standard deviation calculated over five videos per setup. On \textbf{(a)} we show the so-called motion problems and compare against the baselines. On \textbf{(b)} since there are no baseline methods for non-motion scenarios, and thus, we could not evaluate baselines for the LED and free-fall scale videos. Highlighted values indicate the estimates that are closest to the expected values.}
    \label{tb:baselineCompReal}
    \vspace{-0.3cm}
\end{table*}

\vspace{-0.1cm}
\subsection{Real-world video experiments on Delfys75}
\label{sec:realExp}

We evaluate our model and the baselines on Delfys75 dataset and dynamics described in Section~\ref{sec:dataset}. Our model was trained without masks, while PAIG~\citep{jaques2019physics} needs to learn them, and NIRPI~\citep{hofherr2023neural} requires them as input. For both baselines, we used the hyperparameters given in their respective papers. We trained our model for 500 epochs to ensure parameters converged for all experiments; the learning rate was chosen as described in Section~\ref{sec:methods}. Further training details and an analysis of the latent space of our model are given in the supplement. Our model uses a prior $Q\sim \mathcal{N}(0, 1)$ for $\mathcal{L}_2$ (Eq.~\ref{eq:L2}), except for the Torricelli experiment, where the governing equation (Eq.~\ref{eq:torricelli}) includes a square root that introduces conflict if $z < 0$. To avoid this, we change the expected prior to $Q\sim \mathcal{N}(1, 0.2)$. As explained in Section~\ref{sec:methods}, this change leads to a different renormalization factor.

Each model was tested on five videos to compute the mean and standard deviation for each setting. For the pendulum and the LED, the parameter of each setting is estimated. For the sliding block, multiple values of the parameters result in the same dynamics; hence only the total acceleration $a = g(\sin(\alpha) - \mu \cos(\alpha))$ is estimated. The gravitational constant is also predicted in the dropped ball experiment. Baselines are excluded from intensity and scale experiments as they only support motion dynamics. Table~\ref{tb:baselineCompReal} shows the ground truth (GT) and the learned value of each parameter.

On all settings, our proposed model performs parameter estimation relatively accurately compared to baselines. In particular, we observe that PAIG~\cite{jaques2019physics} estimates parameters only close to the initial value (1.0). We train all models on a single video at a time on each setting. This leads to a small training set compared to the 10,000 videos used in~\cite{jaques2019physics}, illustrating the data efficiency of our model compared to PAIG~\cite{jaques2019physics}. On the other hand, parameter estimates for NIRPI~\cite{hofherr2023neural} converge to varying values, but are on average lower accuracy than those fitted by the proposed model.

\vspace{-0.1cm}
\section{Discussion and Limitations}
\label{sec:disnlim}
We present a novel model and dataset for unsupervised physical parameter estimation from video. While previous methods do not study phenomena other than motion, we go beyond motion and include a variety of dynamical systems.

The proposed model avoids frame prediction, which is challenging, but useful for visualization. However, decoder-based methods need to be designed for the particular dynamical system, as a transformation-agnostic decoder may struggle with high-quality reconstruction. Without decoders, our approach works across various dynamical systems.

In addition, we directly examine latent space predictions, unlike prior models~\citep{hofherr2023neural, jaques2019physics, kandukuri2020learning}, which focus on frame reconstruction and do not discuss the accuracy of predicted dynamics. Since parameter estimation is the main goal, reconstruction is simply a tool to define the unsupervised loss, and therefore, the latent space should be analysed closely.


Our model does not resolve the absolute scale of the state variable, for example, that the pendulum goes from $[-180.0^{\circ},180.0^{\circ}]$. Yet, thanks to the loss function, the model solves its own metric, ensuring assumptions made in Section~\ref{sec:LossFunction}. The choice of the prior target distribution simply normalizes the dynamics and should not affect performance, as seen in Section~\ref{sec:realExp} with the Torricelli experiment. Baselines implicitly or explicitly do this normalization using the spatial transformer, forcing the prediction to be in the pixel metric. In contrast, all our dynamics depend on the prior distribution assumed by the KL-divergence.

Baseline methods are sensitive to $\gamma$ initialization. Our method may also converge to local minima with small learning rates and high initial parameters. Therefore, we enforce variance in the learning rate depending on the initial values of the parameters, which allow for wider search spaces.


Our model demonstrates good performance on real-world videos, successfully handling challenges such as shadows and perspective distortions, without the need for object masks. Additionally, our approach outperforms baseline models--which prioritize reconstruction tasks--on parameter learning. Using a spatial transformer, baselines rely on the positions and velocities predicted by the encoder, overlooking minor variations introduced by the governing equation.

Our new real-world dataset is versatile: While baseline models do not support non-motion dynamics and require object masks, they could still be trained on the motion-based subset of our proposed dataset with the accompanying masks. Thus, we believe our dataset will be helpful for evaluating future models on a common dataset, enabling more comprehensive comparisons in complex real-world settings.

\textbf{Limitations.} Our model is suited for continuous, autonomous differential equations. However, some systems, such as fluids, are described with more complex differential equations. In addition, we 
need to guarantee that Eq.~\ref{eq:physblock} is differentiable. Taken together, our model needs to be extended to tackle more complex use cases, including multiple objects in a scene with independent dynamics.

Due to discretization, performance may vary with resolution. For example, small changes of the dynamic variable in the scale dataset were challenging to detect as the model cannot achieve sub-pixel accuracy. 

We avoid predicting object masks in videos. Real-world experiments in Section~\ref{sec:realExp} show that our model does not need masks when trained with complex backgrounds, whereas baseline methods still require them~\citep{jaques2019physics, hofherr2023neural}. However, when multiple objects interact, masks become necessary, as demonstrated in experiment~\ref{sec:baselinesComparision}. While learning masks is an effective solution~\citep{hsieh2018learning, jaques2019physics}, it relies heavily on the spatial transformer limited to roto-translation problems. 

Nevertheless, limitations are mostly related to the early stages of the problem, opening multiple opportunities for future works; with our work, we provide a base which can be extended to more complex problems in different scientific domains and initiate data sharing to evaluate future models.
\vspace{-0.2cm}

\newpage

{
    \small
    \bibliographystyle{ieeenat_fullname}
    \bibliography{main}
}
\newpage

\appendix

\section{Appendix / Supplemental material}
\begin{figure*}[t]
    \centering
    \includegraphics[width=0.9\textwidth]{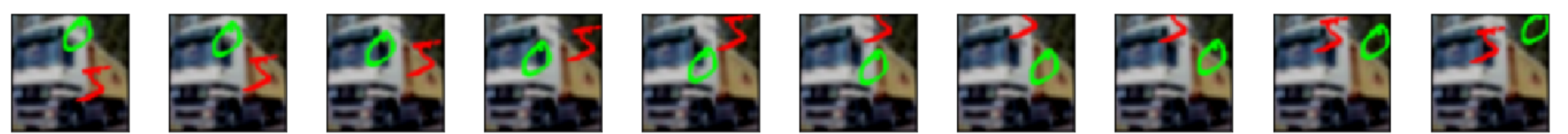}
    \caption{ \textbf{Dataset baseline}. It shows the evolution of the spring dynamical system of two MNIST digits over a static CIFAR10 background. Figure adapted from~\citep{jaques2019physics}.}
    \label{fig:enter-label}
\end{figure*}

\subsection{Synthetic Dataset Dimensionality}\label{sec:detailbaselines}

Each experiment uses an image size of $(50 \times 50)$ pixels, and the simulated dynamics is the same for the three datasets (Eq.~\ref{eq:secondOrder}) normalized with respect to the image size and maximum pixel intensity.
Each dataset consists of 500 training samples with 20 frames, with a final numerical dimensionality of $(samples \times frames \times \# channels \times width \times height) = (500 \times 20 \times 1 \times 50 \times 50)$. 

\subsection{Training Hyperparameters}\label{sec:detailstraining}

The experiments and baselines were executed on an NVIDIA 3080 GPU. For our model, implemented in PyTorch, the encoder was trained using the Adam optimizer~\citep{kingma2014adam} with a learning rate of $1 \times e^{-2}$ and the Kaming weight initialization for MLP layers.

\subsection{Simulation Details}\label{sec:details2ndorder}

Here we discuss the details of the dynamics simulation of the experiments in section~\ref{sec:paramAcc}. The equation ~\ref{eq:secondOrder} represents a harmonic oscillator with the closed solution:

\begin{eqnarray}
    z(t) = A e^{-\zeta t}cos(\omega t + \phi).
    \label{eq:close2ndorder}
\end{eqnarray}

Where $\omega = 2$ is the frequency we used for simulation and $\zeta = 0.04$ the damping factor.  This parameter relates to $\gamma$ as follows:
\vspace{-0.5cm}
\begin{eqnarray}
\label{eq:GT}
    &\gamma_0 = \omega^2 + \zeta^2 = 4.0016.  \\
    &\gamma_1 = 2\zeta= 0.08.
\end{eqnarray}

\subsection{Ablation study}

\begin{figure}[H]
    \centering
    \includegraphics[width=0.35\textwidth]{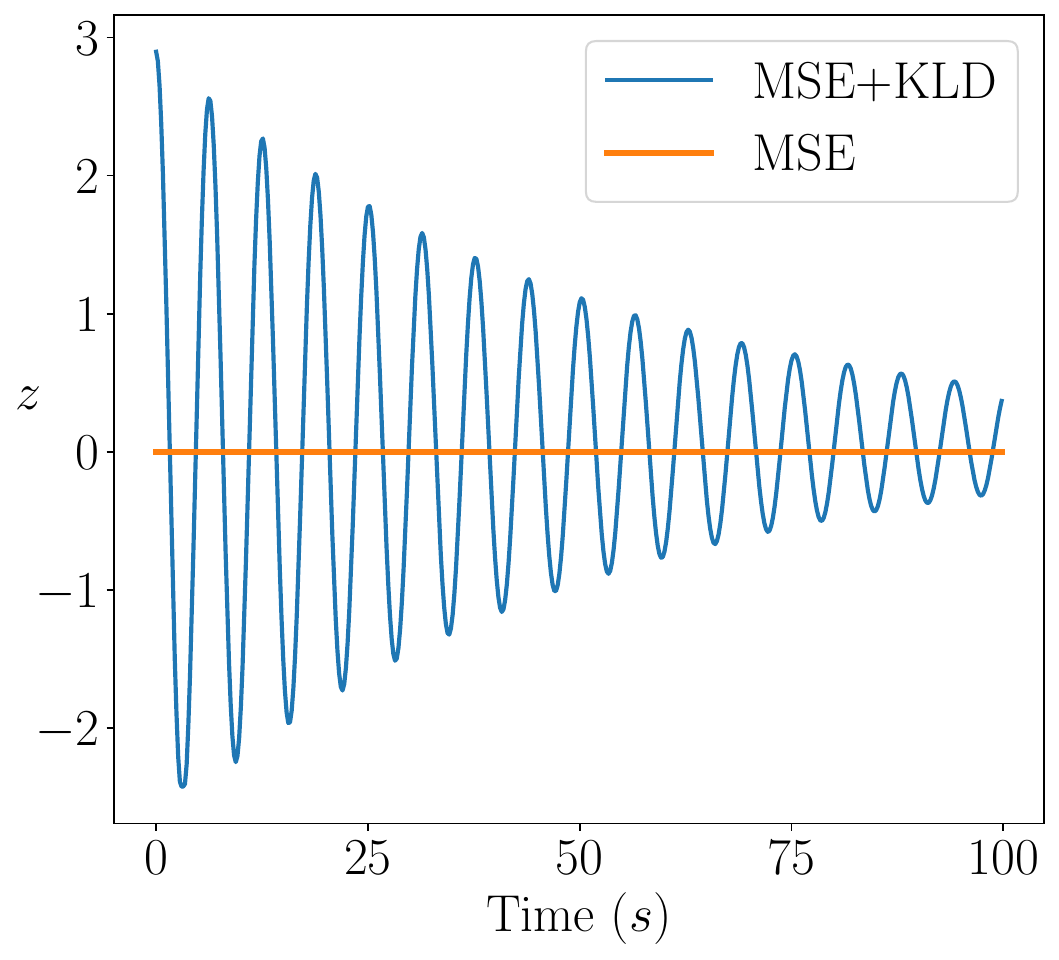}
    \caption{Comparison of the model trained with the MSE loss only (orange) and with our additional KLD loss term. The model relying just on MSE converges to the mean value of the dynamics, while our term guarantees that the model has the expected diversity.}
    \label{fig:Ablation}
\end{figure}

We present an ablation study to show the effect of the  KL-divergence (KLD) term in our loss function; we used the intensity experiment presented in the synthetic experiments section. Table~\ref{tb:Ablation} shows the comparison of learned parameters of the physical equations along with their expected, ground truth (GT) values. In addition, Figure~\ref{fig:Ablation} shows the convergence discussed in the methods section, where a shortcut for the model to optimize the mean squared error (MSE) in the latent space is to converge always to the mean value of the dynamic variable $z$.

\begin{table}[H]
    \centering
    \begin{tabular}{cccc}
        \toprule
        Parameter&MSE+KLD&MSE&\textbf{GT} \\
        \midrule
        $\gamma_0$&3.99& 5.7&4 \\
        $\gamma_1$&0.08& 6.6&0.08\\        
        \bottomrule
    \end{tabular}
    \caption{Ablation comparison of the two-term losses. The model relaying only MSE cannot learn the expected values, while the KLD term allows it to get a proper estimation.}
    \label{tb:Ablation}
\end{table}

\subsection{Baseline Dataset}
\label{sec:baselineData}

In this section, we describe the dataset used to compare the baselines. Both baselines were tested on the dataset published by~\citep{jaques2019physics} and also used in~\citep{hofherr2023neural}. The equation of motion used for both systems is Eq.~\ref{eq:springMnist}
\begin{equation}
    \Vec{F}_{ij}= - k (\Vec{p}_i - \Vec{p}_j ) - l \frac{\Vec{p}_i - \Vec{p}_j}{|\Vec{p}_i - \Vec{p}_j|}.
\label{eq:springMnist}
\end{equation}
where $k$ is the spring constant and $l$ is the equilibrium distance.

\subsection{Delfys parameter estimation}

Below, we provide the details on how the physical parameters and measurement errors were obtained. All parameters with their errors are shown in Table~\ref{tab:estimations_and_errors}.

\begin{table*}[t]
    \centering
    \begin{tabular}{|c|c|c|c|}
        \hline
        \textbf{Scenario} & \textbf{Setting 1} & \textbf{Setting 2} & \textbf{Setting 3} \\
        \hline
        \multirow{2}{*}{Pendulum} & $L = 45 \pm 0.1$ [cm] & $L = 90 \pm 0.1$ [cm] & $L = 150 \pm 0.1$ [cm] \\
        \cline{2-4}
        & $\zeta = -0.059 \pm 0.008$ & $\zeta = -0.059 \pm 0.008$ & $\zeta = -0.059 \pm 0.008$ \\
        \hline
        \multirow{1}{*}{Torricelli} & $k = 0.0095 \pm 10^{-4}$ $\left[ \frac{\sqrt{m}}{s^2} \right]$& $k = 0.0128 \pm 10^{-4}$ $\left[ \frac{\sqrt{m}}{s^2} \right]$ & $k = 0.0162 \pm 2 \cdot 10^{-4}$ $\left[ \frac{\sqrt{m}}{s^2} \right]$\\
        \hline
        \multirow{2}{*}{Sliding block} & $\alpha = 20 \pm 1$ [deg.] & $\alpha = 25 \pm 1$ [deg.] & $\alpha = 30 \pm 1$ [deg.] \\
        \cline{2-4}
        & $\mu = 0.21 \pm 0.04$ & $\mu = 0.21 \pm 0.04$ & $\mu = 0.21 \pm 0.04$ \\        
        \hline
        \multirow{1}{*}{LED} & $\gamma = 2.3$ & $\gamma = 0.92$ & $\gamma = 0.46$\\
        \hline
        \multirow{3}{*}{Free fall} & $r_0 = 3.35 \pm 0.1$ [cm] & $r_0 = 6 \pm 0.1$ [cm] & $r_0 = 10 \pm 0.1$ [cm] \\
        \cline{2-4}
        & $h_0 = 20 \pm 0.1$ [cm] & $h_0 = 20 \pm 0.1$ [cm] & $h_0 = 20 \pm 0.1$ [cm] \\
        \cline{2-4}
        & $f = 1451 \pm 118$ $\left[\frac{\text{pixels}}{\text{m}}\right]$ & $f = 1451 \pm 118$ $\left[\frac{\text{pixels}}{\text{m}}\right]$ & $f = 1451 \pm 118$ $\left[\frac{\text{pixels}}{\text{m}}\right]$\\
        \hline
    \end{tabular}
    \caption{Physical parameters for each experiment and their measurement errors. The first parameter per scenario is changed between settings.}
    \vspace{-0.5cm}
    \label{tab:estimations_and_errors}
\end{table*}

\textbf{Pendulum.} The two parameters to approximate are the length of the pendulum $L$ and the damping coefficient $\zeta$. The length $L$ can be trivially measured, and the estimation error is the spacing between marks on the used tape measure, in this case 0.1~cm. Assuming a small initial angle $\theta_0$, the horizontal offset $x$ of the pendulum can be described by:

\begin{equation}
    x = A\exp\left(-\frac{\zeta}{2}t\right)\cos(\alpha t - \phi).
\end{equation}
with $A$ the amplitude, $\alpha$ the frequency, and $\phi$ the initial phase. The peaks of the above curve then decay as ${x_\text{peak} = A\exp\left(-\frac{\zeta}{2}t\right)}$. The damping coefficient $\zeta$ can therefore be obtained by fitting linear regression to the function $\ln(x_\text{peak})$. This approximation is done on the setting with the longest length of the string, due to the lowest initial angle $\theta_0$. The error corresponds to the standard deviation over the 5 videos in this setting.

\textbf{Torricelli.} The only parameter to estimate in this scenario is $k$ relater to the water flow rate. Given the initial height of the water $h_0$, the final height of the water $h_t$, and the length of the video $t$, the parameter $k$ can be computed as:

\begin{equation}
    k = 2\frac{\sqrt{h_0} - \sqrt{h_t}}{t}.
\end{equation}

The video clips were cut such that initial and final heights were correspondingly always $h_0 = 7$cm and $h_t = 1$cm. The error estimate is obtained by computing the standard deviation of the parameter $k$ over the five videos for each setting.

\textbf{Sliding block.} The two parameters to be measured are the inclination angle $\alpha$ and the friction coefficient $\mu$. The inclination angle was set by varying the height of the top of the ramp. Investigating the recorded videos with a protractor showed that the angle was correct to within one degree. Using the sliding block equation as shown in the paper, the friction coefficient can be computed as:

\begin{equation}
    \mu = \tan(\alpha) - \frac{2s}{gt^2}.
\end{equation}

with $s=72.6$cm the total travel distance of the block and $t$ the duration of the video. Since the friction coefficient should be constant across all settings, the estimate and the error are computed over all 15 videos for this experiment. The error corresponds to the standard deviation.

\textbf{LED.} The only relevant parameter is the decay $\gamma$. This decay is controlled automatically, and thus, the value is exact.

\textbf{Free fall.} The radius $r_0$ is measured with a tape measure with an error of $0.1$cm. The initial distance from the camera $h_0=20$cm was measured likewise. The focal length if calculated as $f = \frac{r(0)}{r_0}h_0$ and the error is the standard deviation of the focal length over 15 recordings for this experiment. It should be noted that the units of the focal length are $\left[\frac{\text{pixels}}{\text{m}}\right]$ as the focal length makes the conversion between metric and pixel spaces. Also for this reason, the focal length cannot be taken from the producer's spec sheet for the used camera.

\subsection{Delfys75 real-video training details}
\label{sec:rvtrain}

For all videos, we adhered to the specifications outlined in the methods section. The model was trained over 500 epochs with an initial learning rate $lr = 1e^{-2}$. Since each experimental group contained a different number of frames due to varying dynamics, the batch size and the number of input frames per sample were adjusted accordingly. The delta time ($dt$), defined as the time interval between frames, is determined by the camera's recording speed in frames per second ($fps  = 60$). While the minimum possible ${dt = \frac{1}{fps}}$  was an option, it was not suitable for all experiments. When frame-to-frame differences were negligible ($x_i \approx x_{i+1} $), $dt$ was increased to ensure meaningful variations between frames for prediction.

\begin{table}[h]
    \centering
    \begin{tabular}{l >{\centering\arraybackslash}m{1.0cm} >{\centering\arraybackslash}m{1.5cm} >{\centering\arraybackslash}m{1.5cm}}
        \toprule
        Experiment & Batch Size & Frames per Sample & $dt \: (s)$ \\
        \midrule
        \addlinespace[1pt]
        Pendulum & 64 & 20 & $\frac{1}{10}$ \\ 
        \addlinespace[1pt]
        Torricelli & 64 & 20 & $\frac{1}{10}$ \\ 
        \addlinespace[1pt]
        Sliding Block & 32 & 10 & $\frac{1}{30}$ \\ 
        \addlinespace[1pt]
        LED & 32 & 20 & $\frac{1}{60}$ \\ 
        \addlinespace[1pt]
        Free Fall Scale & all & 4 & $\frac{1}{30}$ \\ 
        \bottomrule
    \end{tabular}
    \caption{Hyperparameters used for training on Delfys75 experiments.}
    \label{tb:TrainingDetailsReal}
\end{table}

\begin{figure*}[h]
    \centering
    \begin{subfigure}[b]{0.8\textwidth}
        \centering
        \resizebox{0.7\textwidth}{!}{
        \begin{tabular}{clcc}               
            \toprule        
            \textbf{Equation} & \textbf{Parameter} & \textbf{Expected} $(m)$ & \textbf{Estimated} $(m)$ \\ \midrule
     $\theta^{(2)} = -\zeta \theta^{(1)} - \frac{g}{L}\sin(\theta)$ & String length (L)          &                  1.20                      &                    1.22                 \\
            \bottomrule
        \end{tabular}}
        \caption{}
    \end{subfigure}
    \begin{subfigure}[b]{0.8\textwidth}
        \centering
        \includegraphics[width=0.8\textwidth]{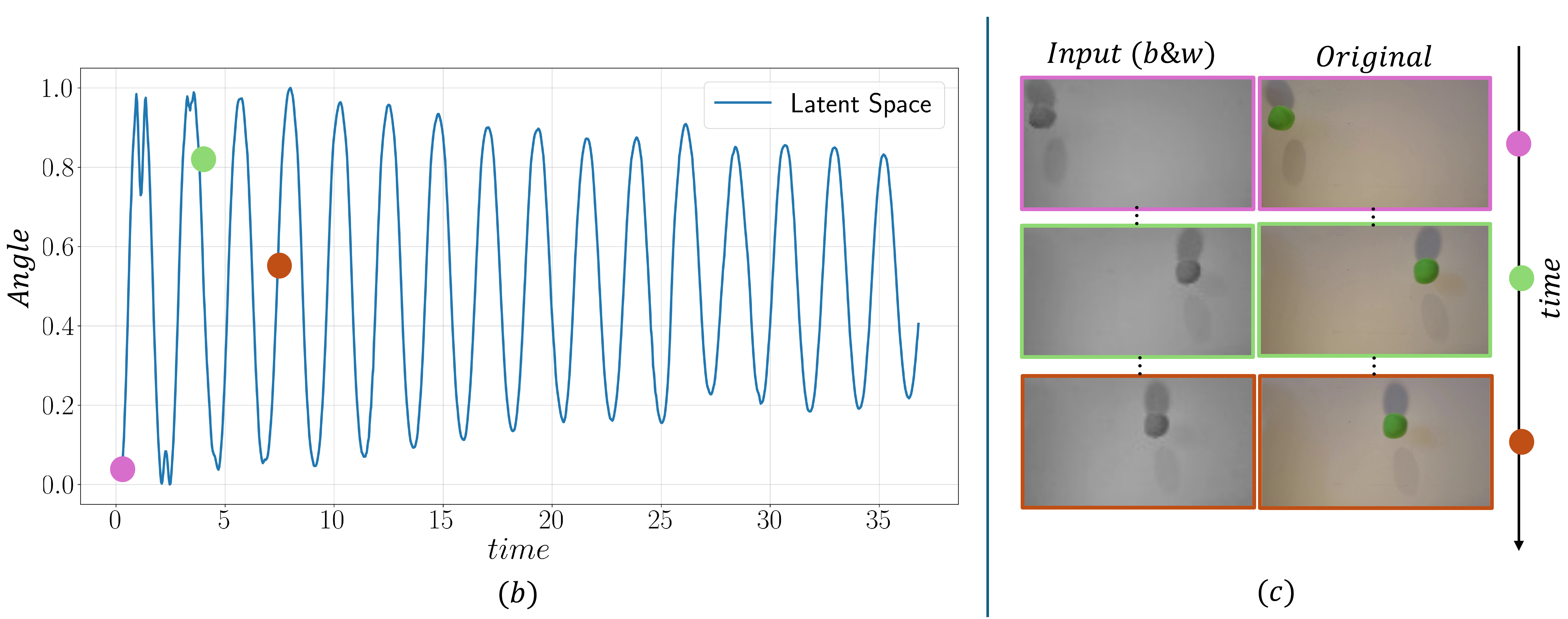}
    \end{subfigure}
    \caption{Real-world pendulum recording parameter estimation. \textbf{(a)}~The angle $\theta$ is the latent variable. Damping factor $\zeta$ and the string length $L$ are learned. \textbf{(b)}~Extracted dynamics by the model. \textbf{(c)}~Gray scale input and the original frame from the dataset, related to time in plot (b) using the coloured dots. Our model can estimate the parameter $L$ with only a 0.02~m error.}
    \label{fig:Exp_ball}
    \vspace{-0.3cm}
\end{figure*}

\subsection{Ablation study $dt$}

For this experiment, we examine how changing the Euler step $dt$ affects the training algorithm. Note that, due to the properties of Euler's method, high $dt$ values yield poor approximations, while very small $dt$ values weaken the gradient on the learnable parameter. The result are shown on table~\ref{tab:dataset_params}.

\begin{table}[h]
    \centering
    
    \begin{tabular}{>{\centering\arraybackslash}m{4.0cm} rr}
        \toprule
        Dataset (Estimated parameter)                            & \( dt \)  & Value\\
        \midrule         
Intensity                   & 0.2                    & 0.08                                \\
(Damping factor)            & 0.4                    & 0.078                               \\
$\gamma$                      & 0.8                    & 0.077                               \\
\midrule
Torricelli                  & 0.1                    & 0.0089                              \\
(Flow rate)                 & 0.2                    & 0.0089                              \\
$k [\frac{\sqrt{m}}{s}]$                          & 0.4                    & 0.011                               \\
\midrule
Pendulum                    & 0.1                    & 0.45                                \\
(String length)             & 0.2                    & 0.51                                \\
$L [m] $                         & 0.4                    & 0.48      \\                        
        \bottomrule
    \end{tabular}
    \caption{Estimated parameter for various $dt$. Results indicate that it has a small impact on predictions.}
    \vspace{-0.6cm}
    \label{tab:dataset_params}
\end{table}

\subsection{Real video latent space visualization}
\label{sec:rvlatent}

\begin{figure}[t]
    \centering
    \includegraphics[width=0.36\textwidth]{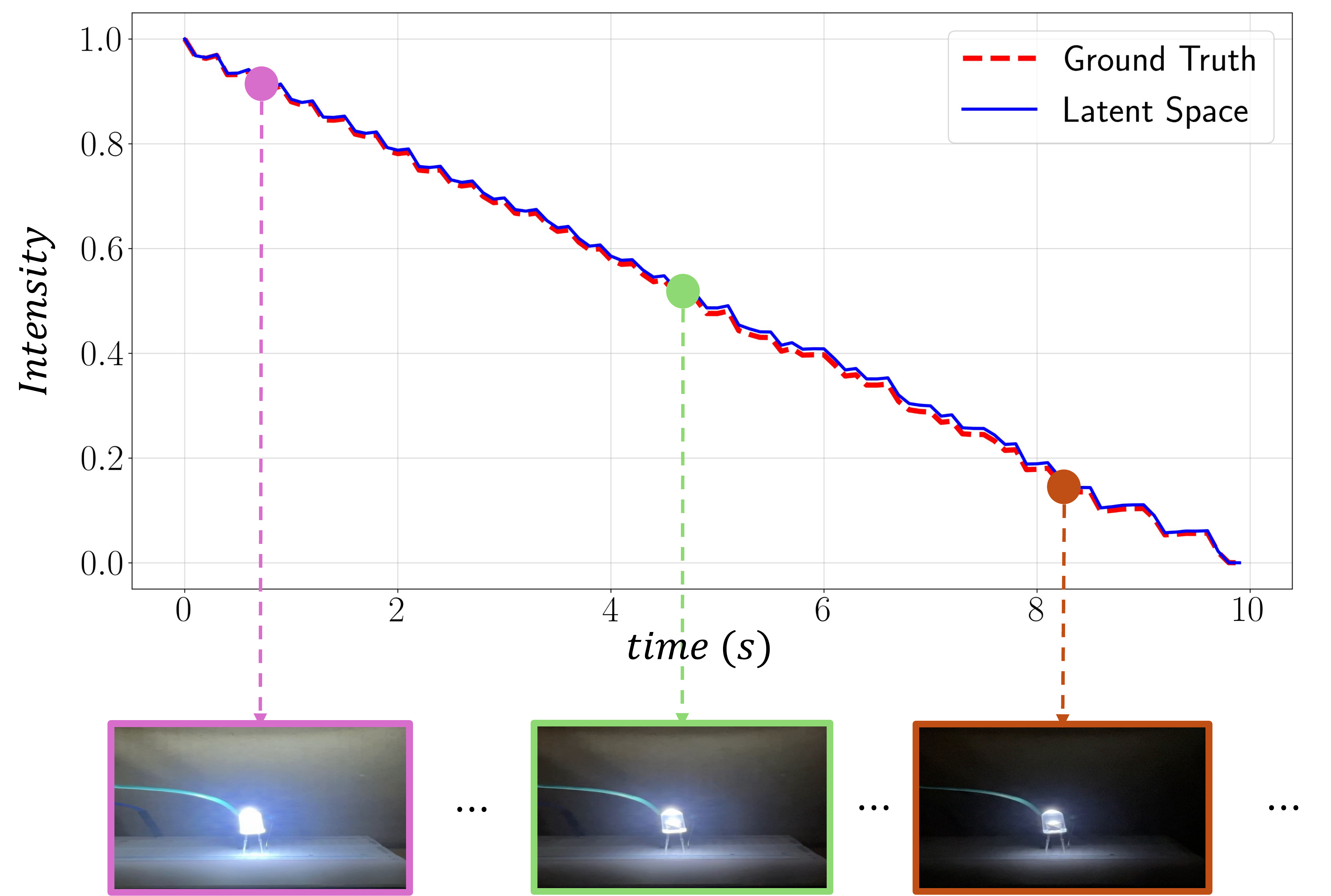}
    \caption{Estimating dynamics from a real-world linear intensity change video. At the top are the estimated dynamics from the model (Latent Space), which are compared with a manually extracted ground truth; the plot shows three dots where example frames of the video inputs are shown in the bottom. The plot shows the latent space can accurately estimate the intensity, capturing the global behaviour over time and following the expected dynamics.}
    \label{fig:Exp_led}
    \vspace{-0.3cm}
\end{figure}

This section presents two different systems: 1. An LED light video recording with the constant brightness change over time~\ref{fig:Exp_led}, similar to the intensity problems previously studied. 2. A pendulum~\ref{fig:Exp_ball} recording which validates the model in a realistic version of the synthetic dataset. For training these models, no masks were needed compared to baselines~\citep{jaques2019physics, hofherr2023neural}.

For an LED recording with constant change of brightness over time,~\ref{fig:Exp_led} shows that the model performs accurately (model predictions in blue) when compared with the ground truth intensity values manually extracted for each frame (red).  In Figure~\ref{fig:Exp_ball}, we present a more realistic use case where we do not have access to the ground truth or precise manual annotations for each frame. Fortunately, the length of the string $L$ is known to be 120 cm (string length is not visible in the video). The task in this experiment is to learn the value of $L$. Quantitatively, in Figure~\ref{fig:Exp_ball}a, the model can reliably estimate the value of the length parameter $L$. Besides, in the latent space Figure~\ref{fig:Exp_ball}b, we can see that the model can accurately predict the natural damped oscillations.


\end{document}